\documentclass[lettersize,journal]{IEEEtran}
\usepackage{amsmath,amsfonts,amssymb,amsthm}
\usepackage{array}
\usepackage{textcomp}
\usepackage{stfloats}
\usepackage{url}
\usepackage{graphicx}
\usepackage{cite}
\usepackage{bm}
\usepackage{bbm}
\usepackage{mathtools}
\usepackage{url}
\usepackage{caption}
\usepackage{makecell}
\usepackage{multirow}
\usepackage{subcaption}
\usepackage{booktabs}

\newcolumntype{C}[1]{>{\centering\let\newline\\\arraybackslash\hspace{0pt}}m{#1}}

\begin{document}


\title{Heterogeneous Graph Sequence Neural Networks for Dynamic Traffic Assignment}

\author{Tong Liu, Hadi Meidani
\thanks{Tong Liu is with the Department of Civil and Environmental Engineering, University of Illinois at Urbana-Champaign.}
\thanks{Hadi Meidani is with the Department of Civil and Environmental Engineering, University of Illinois at Urbana-Champaign.}}

\maketitle

\begin{abstract}
Traffic assignment and traffic flow prediction  provide critical insights for urban planning, traffic management, and the development of intelligent transportation systems. An efficient model for calculating traffic flows over the entire transportation network could provide a more detailed and realistic understanding of traffic dynamics. However, existing traffic prediction approaches, such as those utilizing graph neural networks, are typically limited to locations where sensors are deployed and cannot predict traffic flows beyond sensor locations. To alleviate this limitation, inspired by fundamental relationship that exists between link flows and the origin-destination (OD) travel demands, we proposed the Heterogeneous Spatio-Temporal Graph Sequence Network (HSTGSN).  HSTGSN exploits dependency between origin and destination nodes, even when it is long-range, and learns implicit vehicle route choices under different origin-destination demands. This model is based on a  heterogeneous graph which consists of  road links,  OD links (virtual links connecting origins and destinations) and a spatio-temporal graph encoder-decoder that captures the spatio-temporal relationship between  OD demands and flow distribution. We will show how the graph encoder-decoder is able to recover the incomplete information in the OD demand, by using node embedding from the graph decoder to predict the temporal changes in flow distribution. Using extensive experimental studies on real-world networks with complete/incomplete OD demands, we demonstrate that our method can not only capture the implicit spatio-temporal relationship between link traffic flows and OD demands but also achieve accurate prediction performance and generalization capability.
\end{abstract}


\section{Introduction}
\label{sec:introduction}
\IEEEPARstart{T}{raffic} flow prediction plays a significant role in smart city applications, such as network planning \cite{medina2022urban}, reliability analysis \cite{liu2022graph}, resource management \cite{liu2023optimizing}, etc. Graph neural networks (GNNs) have become instrumental in forecasting traffic patterns within intelligent transportation systems, with notable models including DCRNN \cite{li2017diffusion}, STGCN \cite{yu2017spatio}, GaAN \cite{zhang2018gaan}, DDGCRN \cite{weng2023decomposition}, and STAEformer \cite{liu2023spatio}. These models employ spatio-temporal graph neural networks with traffic data from sensors, where graph nodes represent traffic sensors and edges represent inter-node dependencies. Nevertheless, the effectiveness of these approaches to model the traffic flow over the entire transportation network relies  on how comprehensively the sensors cover the network. This prerequisite is not always satisfied, creating a critical research gap in traffic flow prediction methodologies.

To address this gap, we seek to consider the fundamental relationships that underpin traffic flow distributions in a  transportation system. Specifically, this is the fundamental relationship between traffic flows and vehicle travel plans moving from various origins to their respective destinations in the network, collected in the origin-destination (OD) demands. In this ``physics-based" approach, we exploit the OD demand values as a significant factor impacting the traffic flow pattern.

In addition to regional OD demand,  link flows are also dependent on the collective route choices of individual travelers and network congestion conditions. To account for these factors,  dynamic traffic assignment (DTA)  analyzes link-wise traffic flow patterns by considering the variability in travel agent behaviors and network conditions, thereby facilitating a deeper understanding of traffic distribution across networks \cite{aghamohammadi2020dynamic,medina2022urban}. Various approaches have been proposed to solve the DTA problem, aiming to establish a dynamic user equilibrium (DUE) solution \cite{huang2020dynamic}. The DUE assumes the effective unit travel delay of all travel agents for the same purpose is identical for all departure times \cite{han2003dynamic}. In the context of traffic flow simulation, DUE-based dynamic traffic assignment is used to investigate how travel agents adjust their route choices based on OD demands and network conditions, effectively optimizing traffic flow and enhancing network efficiency.

Since one of the core objectives of DUE-based dynamic traffic assignment is to estimate the link-wise traffic flows, it can be considered as a traffic flow prediction model as well. However, there is a distinct difference between the DUE-based dynamic traffic assignment and the sensor-based traffic flow prediction in the aforementioned literature \cite{li2017diffusion,yu2017spatio,zhang2018gaan,weng2023decomposition,liu2023spatio}. Specifically, the main advantage of equilibrium-based DTA is its capability to predict the traffic flow beyond the coverage of the sensor network, i.e. at locations with no deployed sensor. This shows the broader potential of DUE-based DTA compared to node-based traffic flow prediction methods. Despite significant progress in the applying DTA and DUE on traffic flow modeling, practical  challenges still exist. Notably, the computational demands of executing DTA, particularly in large urban networks, are remarkably high \cite{aghamohammadi2020dynamic}.

To address this computational challenge, in this work, we aim to design an efficient surrogate model for dynamic traffic assignment problems in order to efficiently predict the link-wise traffic flow over a network. Specifically, we develop a spatial-temporal graph neural network model, because of their effectiveness in handling network data. It should be noted that predicting DUE-based traffic flow using GNNs presents several significant challenges. Firstly, the OD demand is a necessary input for the DUE-based DTA problem. However, the message-passing between origin-destination node pairs that are far apart on the network is challenging and as a result long-range dependencies between OD pairs may not be properly accounted using moderate number of message passing steps. Secondly, the dynamic nature of network conditions, such as road closures and traffic incidents, requires frequent re-calculation of the DTA models. These dynamic factors need to also be accurately reflected in the GNN model to reliably predict altered traffic flow distributions. Last but not least, the regional OD demand, which is the most crucial input of DTA, is often over-estimated or under-estimated due to missing information \cite{hussain2021transit}, which leads to inaccurate traffic flow prediction. These limitations negatively affect the performance of  GNN models for DUE-based traffic flow prediction.

\begin{figure}[htb!]
\centering
\begin{minipage}{0.48\linewidth}
\centering
    \subfloat[road link]{\includegraphics[height=1in]{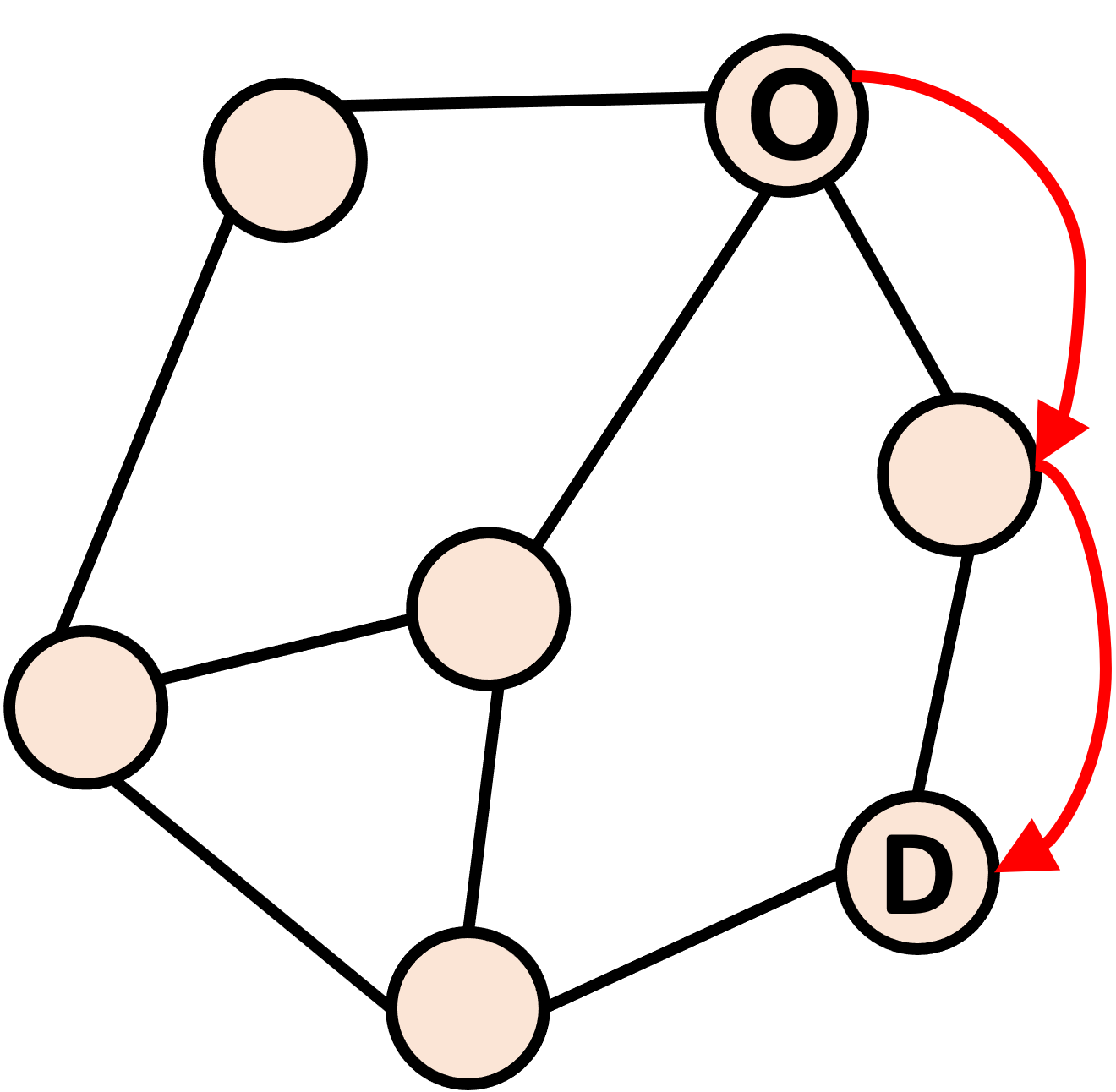}}
    \label{fig:real_link}
\end{minipage}
\begin{minipage}{0.48\linewidth}
\centering
    \subfloat[OD link]{\includegraphics[height=1in]{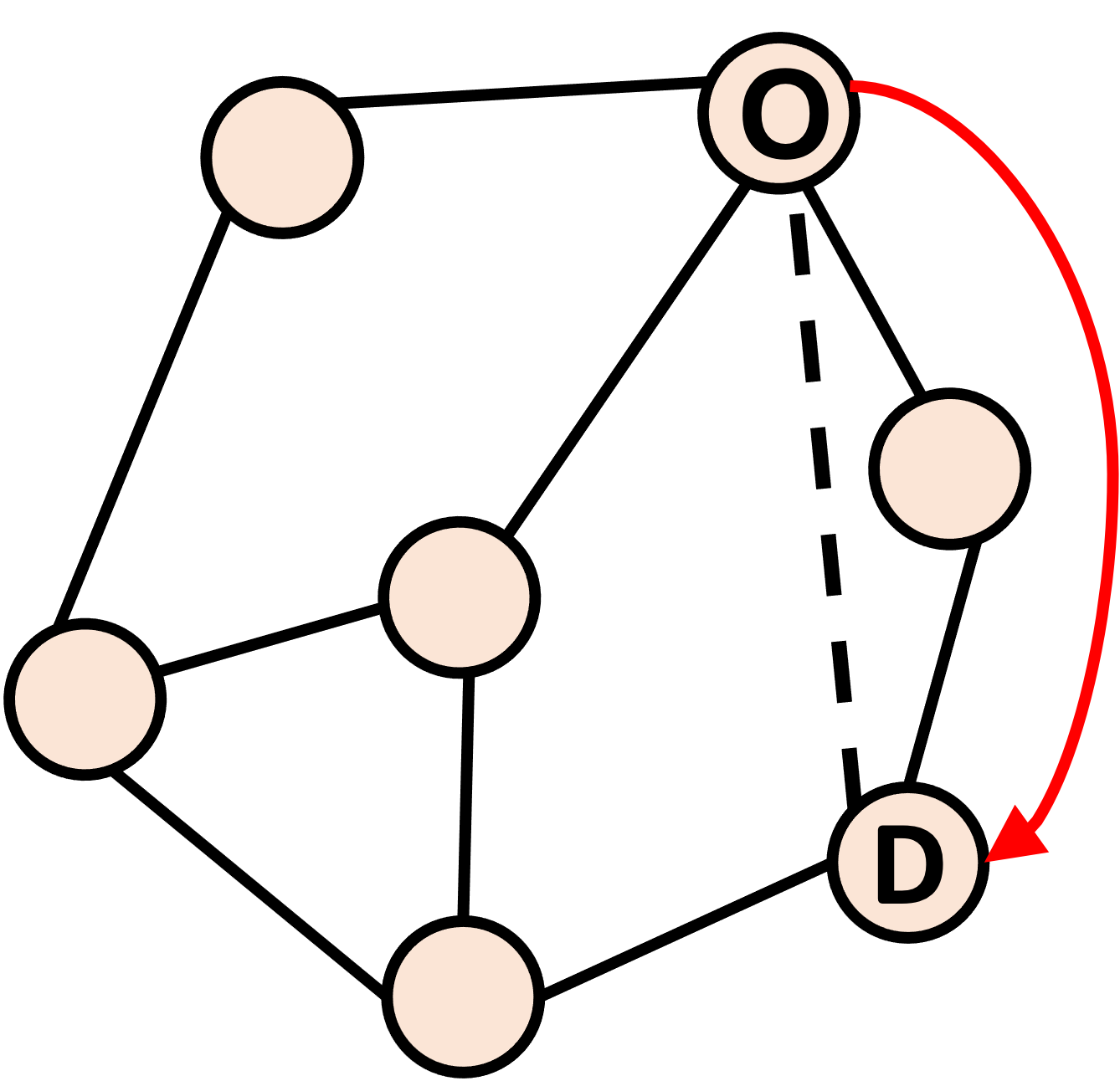}}
    \label{fig:virtual_link}
\end{minipage}
\caption{Illustration of road links (solid lines) and OD links (dash lines). The information exchange between Origin and Destination on road links require two message passing steps (shown by the red curved arrows). When the OD link is added, this exchange is achieved after only one message passing step.}
\label{fig:real_virtual_link}
\end{figure}

To overcome the aforementioned obstacles, in this work, we proposed a novel \textbf{H}eterogeneous \textbf{S}patio-\textbf{T}emporal \textbf{G}raph \textbf{S}equence \textbf{N}etwork (HSTGSN), specifically for DUE-based traffic assignment to predict link-wise traffic flows. The architecture of HSTGSN leverages the recurrent encoder-decoder framework for sequential forecasting. In particular, we construct a heterogeneous transportation graph including two types of links: (1) road links, and (2) OD links. Each set of links represents a distinct topological connection of the transportation system, as illustrated in Figure \ref{fig:real_virtual_link}. Unlike the road links representing the real roadways, OD links are auxiliary links directly connecting  OD node pairs. With the help of these OD links, we can reduce the required number of message-passing steps between OD node pairs to one, thus significantly enhance the computational efficiency. Then, we introduce a spatial-temporal graph encoder-decoder for feature propagation. The node feature is first encoded by an adaptive spatial graph encoder, which could capture the long-range dependency as well as the local topological relationship between nodes. Subsequently, a temporal graph decoder is employed to refine the edge embeddings and predict the temporal flow distribution across the network. Using numerical  experiments on three real-world urban transportation networks, we show the accuracy and computational efficiency of the model in DUE-based traffic assignment problems. By varying OD demands and link conditions, we demonstrate the feasibility and the generalization capability of HSTGSN in learning and predicting dynamic traffic flow patterns. The proposed HSTGSN is also capable of inferring  link traffic flows from  incomplete OD demands. Our contributions can be summarized as follows:


\begin{itemize}
    \item To the best of our knowledge, this is the  first work based on graph neural networks that addresses the critical gap of traffic flow prediction problem left by the absence of extensive sensor network coverage.
    \item We introduce a novel heterogeneous graph encoder-decoder architecture, which comprehensively accounts for both network topologies and OD demand structure. The model is able to capture spatio-temporal relationships between the travel demands, inherent route choices, as well as the local contextual information.
    \item We perform experiments on real-world transportation networks, demonstrating HSTGSN's robust capability in learning and forecasting dynamic traffic flow when complete or incomplete information is available on OD demands.
\end{itemize}



The remainder of this article is structured as follows. General backgrounds on the dynamic user equilibrium and the DUE-based dynamic traffic assignment for traffic flow prediction are presented in Section \ref{sec:preliminaries}. Section \ref{sec:architecture} introduces the detailed architecture of the heterogeneous spatial-temporal graph sequence network for DUE-based traffic flow prediction, and Section \ref{sec:experiment} includes  the numerical results on multiple urban transportation networks. Finally, the conclusions   are presented in Section \ref{sec:conslusion}.

\section{Preliminaries}
\label{sec:preliminaries}
In this section, we first introduce the concept of dynamic user equilibrium and the notations used in this paper. Then, we formalize the problem of DUE-based dynamic traffic assignment.

\subsection{Dynamic User Equalibrium}
Dynamic user equilibrium is a state in network flow problems in which  all drivers have optimized their route choices under the given conditions, such that no individual can reduce their travel time by choosing an alternative route. This concept is supported by the Lighthill-Whitham-Richards (LWR) model \cite{lighthill1955kinematic,richards1956shock}, which approaches macroscopic flow propagation through the lens of fluid dynamics. The LWR model, essential for understanding DUE, introduces a critical relationship between traffic flow and density. This relationship is depicted through a fundamental diagram, which is pivotal in capturing the interaction between traffic density and flow rate. By integrating this relationship into a non-linear, first-order partial equation of density, the LWR model provides a robust theoretical framework to explore and explain the dynamics of traffic, thus facilitating the realization of DUE in practical scenarios \cite{leclercq2007hybrid}. 

Instead of solving LWR model directly, the cell transmission model  \cite{daganzo1994cell} and the link transmission model \cite{yperman2005link} adopt a time and space discretization numerical scheme from the perspective of cell-based dynamics and network link-based dynamics, respectively. These models provide foundational frameworks for understanding and predicting traffic behavior and are instrumental in applying DUE principles to real-world scenarios, helping traffic planners manage congestion and improve overall traffic flow efficiency.

\begin{figure*}[thb!]
\centering
\includegraphics[width=0.95\textwidth]{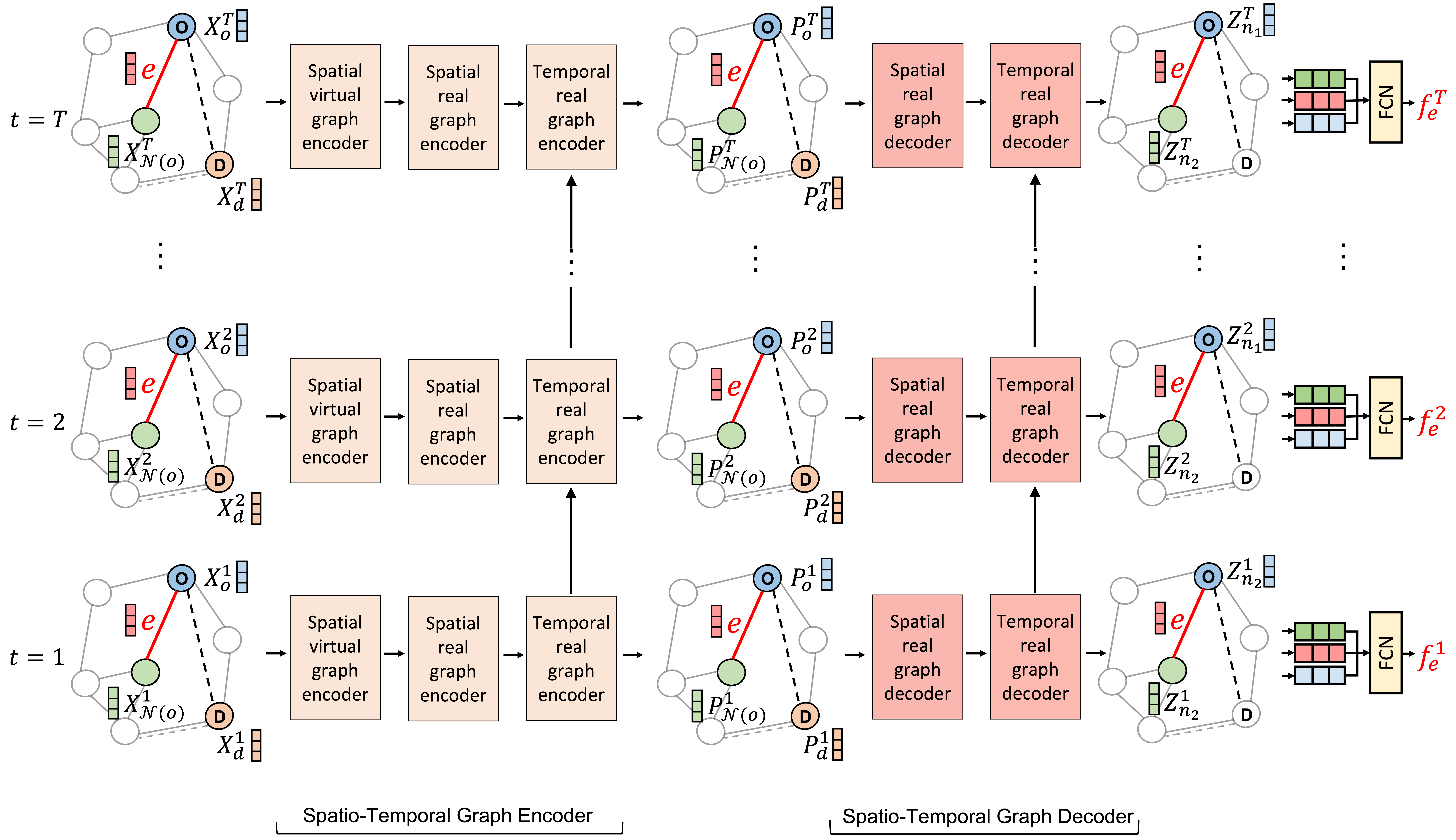}
\caption{Illustration of the heterogeneous spatio-temporal graph sequence network. The solid  and dashed lines in the graphs show road and OD links, respectively. The main component of HSTGSN is the spatio-temporal encoder-decoder. It includes five distinct components: the spatial virtual graph encoder (S-VGE), the spatial real graph encoder (S-RGE), and the temporal real graph encoder (T-RGE), the spatial real graph decoder (S-RGD), and the temporal real graph decoder (T-RGD).}
\label{fig:pipeline}
\end{figure*}

\subsection{DUE-based Dynamic Traffic Assignment}


A transportation network can be represented by $\mathcal{G} = (\mathcal{V}, \mathcal{E})$, where $\mathcal{E}$ and the $\mathcal{V}$ represent the  roads and road intersections, respectively. As mentioned in Section \ref{sec:introduction}, the input of DTA includes temporal OD demands (as node features) and network properties (link features). In this paper, we denote the node-level OD demand as $\bm{X}_n^{1:T} = \{D^{1}, D^{2}, \dots, D^{T}\} \in \mathbb{R}^{|\mathcal{V}| \times N_{\mathcal{V}} \times T}$, where $N_{\mathcal{V}}$ and $T$ are the dimension of the node features and number of time steps, respectively. Furthermore, the link features are denoted with $\bm{X}_e^{1:T} \in \mathbb{R}^{|\mathcal{E}| \times N_{\mathcal{E}} \times T}$, where $N_{\mathcal{E}}$ is the dimension of the link features. 

For a transportation graph $\mathcal{G}$, with their node features $\bm{X}_n^{1:T}$ and link features  $\bm{X}_e^{1:T}$, DUE-based dynamic traffic assignment  seeks to learn a forecasting model $h_{\theta}$ that calculates the traffic flow on all the links:

\begin{equation}
    \tilde{\bm{F}}^{1:T} = h_{\theta}\left(\mathcal{G}, \bm{X}_e^{1:T}, \bm{X}_n^{1:T} ; \theta \right),
\end{equation}
such that the following error is minimized:

\begin{equation}
    L = \sum_{e \in \mathcal{E}}\sum_{t \in T}L_{err}(\tilde{f}_e^{t}, f_e^{t}),
\end{equation}
where $\tilde{\bm{F}}^{1:T} = \{\tilde{f}_e^{t} : e \in \mathcal{E}, t = 1, \dots, T\}$ is the vector of temporal traffic flows on all the links;  $\tilde{f}_e^{t}$ and $f_e^{t}$ are the predicted and ground truth  traffic flow on link $e$ at time step $t$; and $L_{err}$ is an error function, such as squared or absolute errors.

\section{Proposed Method}
\label{sec:architecture}
In this section, we introduce the proposed architecture of HSTGSN for DUE-based dynamic traffic assignment, as shown in Figure \ref{fig:pipeline}. The HSTGSN leverages the recurrent encoder-decoder framework for sequential modeling. The proposed architecture consists of three modules. The first module is to build the heterogeneous graph by introducing auxiliary OD links. The second module propagates the original node and edge features using a spatial-temporal graph encoder. Finally, the third module decodes the graph node embeddings with a spatial-temporal graph decoder on the edge level and predict the dynamic traffic flows. Each of these modules are explained in details in the following sections.

\subsection{Heterogeneous Graph Representation}
Prior works on traffic flow prediction, such as \cite{li2017diffusion,song2020spatial,fang2021spatial}, usually characterize node relationships based on proximity and functional similarity. However, as discussed in Section \ref{sec:introduction},  spatial dependency between the origin and destination nodes is often neglected in these models. To account for this dependency, we add OD links $\mathcal{E}_v$ into the graph model $\mathcal{G}$. The new graph, as depicted in Figure \ref{fig:real_virtual_link},  now includes the real roadway  links $\mathcal{E}_r$, and the virtual OD links, denoted by $\mathcal{E}_v$.  The new heterogeneous transportation graph is now denoted by $\mathcal{G} = (\mathcal{V}, [\mathcal{E}(]\mathcal{E}_r, \mathcal{E}_v])$ with associated node and edge features $\bm{X}_n^{1:T}$ and $\bm{X}_e^{1:T}$. 

Each row $\bm{x}^t_u \in \mathbb{R}^{N_{\mathcal{V}}}$ of the node feature $\bm{X}_n^{1:T}$ corresponds to the feature vector of a single node $u \in \mathcal{V}$ at time step $t$. Each node feature vector includes the origin-destination demand as well as geographical coordinates. Each row $\bm{x}^t_e \in \mathbb{R}^{N_{\mathcal{E}_r}}$ of the real roadway link features $\bm{X}_e^{1:T}$ corresponds to the feature vector of a single edge $e \in \mathcal{E}_r$ at time step $t$. Each edge feature vector includes free-flow travel time and the link capacity. It should be noted that for virtual OD links no intrinsic feature is assigned. Rather, for these OD links, we consider a set of learnable features to capture the dynamics between OD node pairs, which will be further elaborated in the following sections.

An imperative step before graph encoding involves addressing inherent sparsity and normalization deficiencies within the original features. These characteristics can hinder efficient feature propagation within a graph neural network. To overcome this challenge, we incorporate a preprocessing phase to transform the raw features into a low-dimensional representation. Specifically, the raw node features $\bm{X}^{1:T}_n$ is transformed via a multi-layer perception (MLP):

\begin{equation}
    \bm{X}^{1:T} = \mathtt{MLP}(\bm{X}^{1:T}_n; \bm{W}_0, \bm{b}_0),
\end{equation}
where $\bm{X}^{1:T} \in \mathbb{R}^{|\mathcal{V}|\times N_d \times T}$ represent the node embedding including all time steps. And $\bm{X}^{t} \in \mathbb{R}^{|\mathcal{V}|\times N_d}$ represents the node embedding for time step $t$ and $N_d$ denotes the dimension of the node embedding.

\subsection{Spatio-Temporal Graph Encoder}
\begin{figure}[htb!]
\centering
\includegraphics[width=\linewidth]{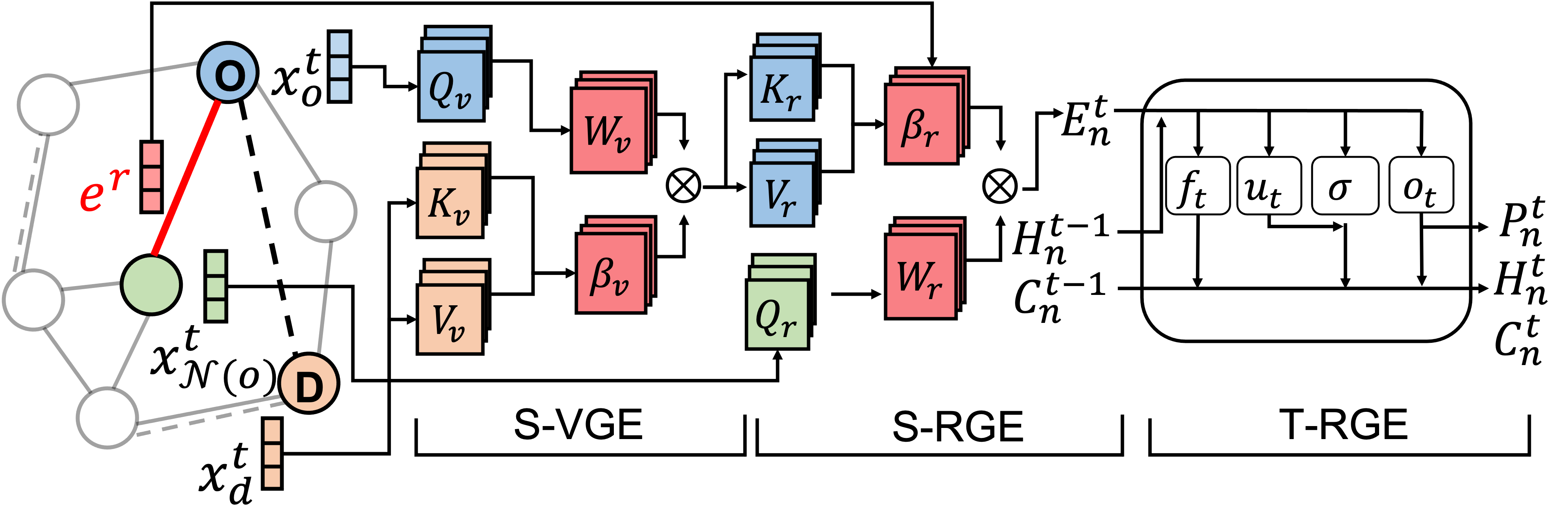}
\caption{Illustration of spatial-temporal graph encoder}
\label{fig:encoder}
\end{figure}

The spatial-temporal graph encoder, as shown in Figure \ref{fig:encoder}, is assembled in a sequential way. The spatio-temporal graph encoder comprises three distinct components: the spatial virtual graph encoder (S-VGE), the spatial real graph encoder (S-RGE), and the temporal real graph encoder (T-RGE). In following sections, we will elaborate on the functionality of each component.

\subsubsection{Spatial Virtual Graph Encoder (S-VGE)}
The main objective of S-VGE is to propagate OD demand features and capture the long-range dependency between origin-destination node pairs through the OD link. OD links provide a direct path between origin-destination node pairs, which can directly  capture long-range dependencies \cite{liu2024end}. However, the heterogeneous distribution of OD demands across different nodes poses significant obstacles in the training process. For example, the OD demand between one OD pair could differ significantly between peak hour and non-peak hour.  To address this issue, we introduce an adaptive attention mechanism within the S-VGE, which is inspired by the heterogeneous graph transformer \cite{hu2020heterogeneous}. More specifically, we parameterize the weight matrices of the interaction operators into an origin node projection, an edge projection, and a destination node projection. Additionally, we introduce an adaptive edge importance vector to quantify the importance of the OD link. Mathematically, the adaptive attention mechanism of $L^{\mathrm{th}}$ layer can be expressed as:

\begin{equation}
\label{eq:virtual_attention}
\begin{aligned}
\bm{Q}^t_{L,i}, \bm{K}^t_{L,i}, \bm{V}^t_{L,i}  = \bm{X}^{t,L}_{o} \bm{W}_{q,i}, \bm{X}^{t,L}_{d} \bm{W}_{k,i}, \bm{X}^{t,L}_{d} \bm{W}_{v,i}, \\
\tilde{\bm{X}}^t_{L+1,i} =  \mathrm{Softmax} \left(\frac{\bm{Q}^t_{L,i}\bm{W}^{ATT}_{L,i}\bm{K}^t_{L,i}}{\sqrt{d_L}} \odot \beta_{L,i} \right)  \bm{V}^t_{L,i} ,
\end{aligned}
\end{equation}
where $\bm{X}^{(\cdot)}_{o}$ and $\bm{X}^{(\cdot)}_{d}$ represent the node feature matrix of the origin nodes and the destination nodes, respectively; $\bm{Q}_{(\cdot)} \in \mathbb{R}^{|\mathcal{V}|\times N_Q}$, $\bm{K}_{(\cdot)} \in \mathbb{R}^{|\mathcal{V}|\times N_Q}$, and $\bm{V}_{(\cdot)} \in \mathbb{R}^{|\mathcal{V}|\times N_Q}$ is the query, key, and value matrices at step $t$ and $i^{\mathrm{th}}$ head of S-VGE; $\bm{W}_{(\cdot)}$ and $\beta_{(\cdot)}$ are the learnable parameters, and $\odot$ is the Hadamard product. Then, using a position-wise feed-forward layer with residue connection and layer normalization, the output of S-VGE is calculated as follows:

\begin{equation}
\label{eq:virtual_embedding}
\begin{aligned}
\bm{X}^t_{L+1,i} & = \tilde{\bm{X}}^t_{L+1,i} + \mathrm{LayerNorm} \left(\mathtt{MLP} \left(\tilde{\bm{X}}^t_{L+1,i}\right)\right), \\
\bm{X}^t_{L+1} & = \left(\bm{X}^t_{L+1,0} \ \| \bm{X}^t_{L+1,1}\| \dots \| \bm{X}^t_{L+1,N_h} \right),
\end{aligned}
\end{equation}
where $N_h$ is the number of heads in the attention. The adaptive attention scores quantify the relative influence among different origin-destination node pairs. Given that route selections and flow distribution may vary in response to different OD demands, it is essential to adjust attention scores adaptively. This flexibility ensures that the model's attention mechanism remains responsive to the fluctuating nature of traffic patterns, thereby enhancing its predictive accuracy in real-world scenarios.

\subsubsection{Spatial Real Graph Encoder (S-RGE)}
The S-RGE is designed to enhance the capabilities of S-VGE and calculate the impact of OD demands on a local region by accounting the connectivity via real roadway links. In particular, a notable challenge within the S-VGE framework is its inability to update certain nodes that lack OD link connections, leaving these nodes not updated in S-VGE. To address this challenge, these nodes are updated in the S-RGE using the other updated nodes that are connected by road links, ensuring a comprehensive evaluation of direct and indirect node relationships. S-RGE and S-VGE share a similar architecture but slightly differ in the attention score calculation. Mathematically, the adaptive attention mechanism at $M^{\mathrm{th}}$ layer of S-RGE can be written as

\begin{equation}
\label{eq:real_attention}
\begin{aligned}
\bm{Q}^t_{M,j}, \bm{K}^t_{M,j}, \bm{V}^t_{M,j} = \bm{X}^{t, M}_{o} \bm{W}_{q,j}, \bm{X}^{t, M}_{\mathcal{N}(o)} \bm{W}_{k,j}, \bm{X}^{t, M}_{\mathcal{N}(o)} \bm{W}_{v,j}, \\
\tilde{\bm{X}}^t_{M+1,j}=  \mathrm{Softmax} \left(\sum_{p=1}^{P}\frac{\bm{Q}^t_{M,j}\bm{W}^{ATT}_{M,j}\bm{K}^t_{M,j}}{\sqrt{d_L}} \odot \beta_{r,p} \right) \bm{V}^t_{M,j},
\end{aligned}
\end{equation}
where $\bm{X}^{(\cdot)}_{o}$ and $\bm{X}^{(\cdot)}_{\mathcal{N(o)}}$ represent the node feature matrix of the origin nodes and the neighbor nodes, respectively; $\bm{Q}_{(\cdot)} \in \mathbb{R}^{|\mathcal{V}|\times N_Q}$, $\bm{K}_{(\cdot)} \in \mathbb{R}^{|\mathcal{V}|\times N_Q}$, and $\bm{V}_{(\cdot)} \in \mathbb{R}^{|\mathcal{V}|\times N_Q}$ are the query, key, and value matrices at step $t$ and $j^{\mathrm{th}}$ head of S-RGE; and $\bm{W}_{(\cdot)}$ is the learnable parameter and $\beta_{r,p}$ represents the $p^{\mathrm{th}}$ normalized road link feature. 

It should be noted that the node features are propagated in two distinct patterns in S-VGE and S-RGE. Specifically, the S-VGE captures  long-range dependencies between nodes and facilitates the integration of contextual information from non-adjacent nodes. On the other hand, S-RGE ensures that the local topological relationships and interactions between nodes are effectively incorporated. After calculating the adaptive attention score,  similarly to Equation \ref{eq:virtual_embedding}, via a position-wise feed-forward layer and layer normalization,  the output of S-RGE is calculated. For the sake of simplicity, the output from the last layer of S-RGE is denoted as $\bm{E}^t_{n}$, which is used as the input to the temporal graph encoder.

\subsubsection{Temporal Real-Graph Encoder (T-RGE)}
Unlike the spatial encoders handling spatial dependencies, T-RGE focuses on propagating information through the temporal dimension. This module is inspired by sequence-to-sequence models that are developed for time-series data \cite{cai2020graph}.  T-RGE leverages a graph-based LSTM model to encode temporal dynamics. In particular, this  module utilizes a stack of recurrent layers to process sequences of node features, enabling the  integration of temporal patterns across different time steps. This temporal encoding is formulated as follows:

\begin{equation}
\label{eq:lstm}
\begin{aligned}
\bm{g}^{t} &= \sigma(\bm{W}_g \cdot [\bm{H}^t_{n}, \bm{E}^t_{n}] + \bm{b}_g), \\
\bm{i}_t &= \sigma(\bm{W}_i \cdot [\bm{H}^t_{n}, \bm{E}^t_{n}] + \bm{b}_i), \\
\tilde{\bm{C}}^{t}_n &= \tanh(\bm{W}_C \cdot [\bm{H}^t_{n}, \bm{E}^t_{n}] + \bm{b}_C), \\
\bm{C}_n^{t} &= \bm{g}_t * \bm{C}_{t-1} + \bm{i}_t * \tilde{\bm{C}}^{t}_n, \\
\bm{P}_n^{t} &= \sigma(\bm{W}_P \cdot [\bm{H}^t_{n}, \bm{E}^t_{n}] + \bm{b}_P), \\
\bm{E}^{t+1}_{n} &= \bm{P}_n^{t} * \tanh(\bm{C}_n^{t}),
\end{aligned}
\end{equation}
where $\bm{P}^t_{n}$ is the vector of encoded graph features at time step $t$; $\bm{W}_{(\cdot)}$ represents learnable weight matrices; $\bm{H}^{t}_n$ and $\bm{C}_n^{t}$ represent the graph hidden state and cell state at time step $t$. Finally,  T-RGE yields encoded representations $\bm{P}_n^{t}$ that encapsulate the spatio-temporal information for each node at all time steps.


\subsection{Spatio-Temporal Graph Decoder}
\begin{figure}[htb!]
\centering
\includegraphics[width=0.9\linewidth]{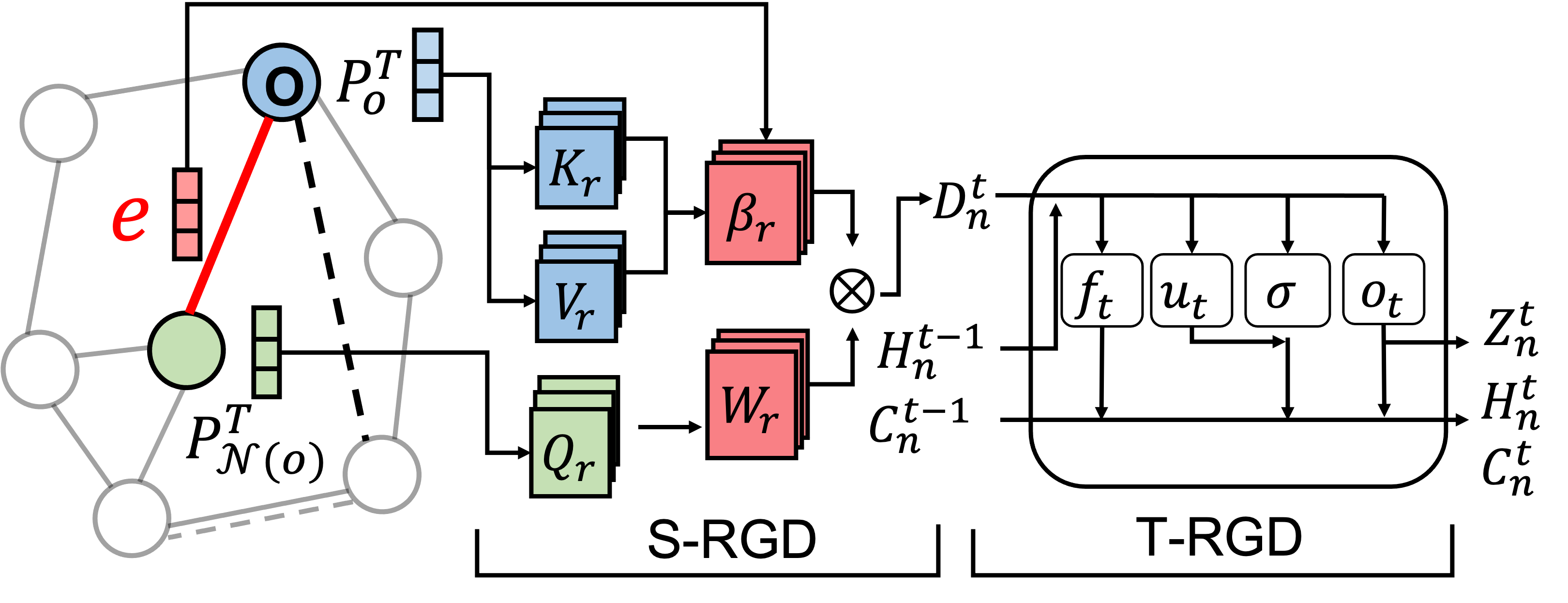}
\caption{Illustration of spatial-temporal graph decoder}
\label{fig:decoder}
\end{figure}

The spatial-temporal graph encoder, shown in Figure \ref{fig:decoder}, has two components: (1) the spatial real graph decoder (S-RGD) and the temporal real graph decoder (T-RGD). The architecture of the spatio-temporal decoder is same as that of the encoder (S-RGE and T-RGE), promoting symmetry in the encoding-decoding pipeline. The encoded node features are first passed into S-RGD and then T-RGD to reconstruct spatio-temporal attributes. A simple  mathematical formulation of spatio-temporal graph decoder can be given by 

\begin{equation}
\label{eq:s-rgd}
\begin{aligned}
\bm{D}^{1:T}_{n} & = \texttt{S-RGD}(\bm{P}^{1:T}_{n}; W_{S}),\\
\bm{Z}^{1:T}_{n} & = \texttt{T-RGD}(\bm{D}^{1:T}_{n}; W_{T}),
\end{aligned}
\end{equation}
where $\bm{D}^{1:T}_{n}$ and $\bm{Z}^{1:T}_{n}$ are the node embedding from the spatial decoder and temporal decoder, respectively.

\subsection{Link-level Traffic Flow Prediction}
The last step of HSTGSN is to predict the link-wise traffic flow. Instead of directly utilizing $\bm{Z}^{1:T}_{n}$ from the spatio-temporal decoder for prediction, we first concatenate node embeddings from endpoints of each edge as well as the edge features to build the edge embedding. We use this  edge embedding as the input to a fully connected network which predicts (1) the temporal traffic flows, and (2)  the link utilization percentages, i.e.,

\begin{equation}
\label{eq:edge_prediction}
\begin{aligned}
\bm{S}^{t}_{e} &= [\bm{Z}^{t}_{n_1} \| \bm{Z}^{1:T}_{n_2} \| \bm{X}^{1:T}_e], \quad \forall e=(n_1, n_2) \in \mathcal{E}_r, \\
[\tilde{f}^{t}_{e}, \tilde{\alpha}^{t}_{e}] &= \texttt{MLP}(\bm{S}^{t}_{e}; \bm{W}_e, \bm{b}_e), \quad t= 1, 2, \dots, T,
\end{aligned}
\end{equation}
where $\tilde{f}^{t}_{e}$ and $\tilde{\alpha}^{t}_{e}$ are the predicted link flow and link utilization percentage on link $e$, respectively. The proposed model utilizes $L_1$ loss as the loss function, which measures the discrepancy between prediction and ground truth according to
\begin{equation}
    L_s = \frac{1}{|\mathcal{E}_r|\times |T|} \sum_{t \in T} \sum_{e \in \mathcal{E}_r}|f_e^{t} - \tilde{f}_e^{t}|,
    \label{eq:ls}
\end{equation}
where the $f_e$ and $\tilde{f_e}$ represent the ground truth and predicted  flow on link $e \in \mathcal{E}$.

\section{Experiments}
\label{sec:experiment}
In this section, we numerically evaluate the performance of HSTGSN within the context of DUE-based traffic assignment for traffic flow prediction. Our evaluations aim to answer the following questions:
\begin{itemize}
    \item How does HSTGSN compare against  other spatial-temporal graph neural network models for traffic flow prediction under dynamic user equilibrium?
    \item How is the generalization performance of HSTGSN under the scenarios of incomplete origin-destination demand?
    \item How does each component of the HSTGSN model contribute to the performance of the proposed model?
\end{itemize}

\subsection{Experiment Setup}
\subsubsection{Dataset}
We conducted multiple numerical experiments on three real-world urban transportation networks, namely Sioux Falls, Anaheim, and Chicago network. The network information, including topologies and associated characteristics and origin-destination demand, is obtained from \cite{bar2021transportation}. The detailed information about these networks are shown in Table \ref{tab:urban_network_detail}. 

\begin{table}[htb!]
\centering
\caption{The characteristics of the studied networks.}
\label{tab:urban_network_detail}
\begin{tabular}{ccccc}
\hline
Network Name & $|\mathcal{V}|$ & $|\mathcal{E}|$ & Average Degree & Total Demand (veh.) \\ \hline
Sioux Falls  & 24      & 76      & 3.17           & 188,960          \\
Anaheim      & 416     & 914    & 3.05           & 226,279          \\ 
Chicago          & 933      & 2560     & 3.88           & 980,226          \\ 
\hline
\end{tabular}%
\end{table}

We choose DTALite as the ground truth solver for dynamic traffic assignment experiments \cite{zhou2014dtalite}. The platform leverages a mesoscopic traffic flow simulation simulation, enabling high-resolution modeling of individual vehicle movements and overall traffic dynamics. The duration and resolution for the DTA simulation is 8 hours and 1 minute, respectively. The link flows across the network are simulated every one minute, which are subsequently aggregated in 5-minute intervals, resulting in a total of 97 time steps. 

To increase the sample variations and validate the generalization capability of HSTGSN, we scaled the individual origin-destination demand for all the time steps by a scaling factor according to
\begin{equation}
\label{eq:od_demand}
    \tilde{D}^{t}_{i,j} = \delta^{\text{OD}}_{i,j,t} \ D^{t}_{i,j},
\end{equation}
where $D_{i,j}$ is the default OD demand between nodes $i$ and $j$, and $\delta^{t}_{i,j} \sim U(0.5, 1.5)$ is a uniformly distributed random scaling factor, independently drawn for  OD pair ($i$, $j$) at time $t$. Additionally, we seek to also account for changes in network properties caused  by traffic accidents, road construction, road damages, or adverse weather conditions. To induce variations in network properties, random reductions in  link capacities are considered according to
\begin{equation}
\label{eq:ca_demand}
    \tilde{c}_{a} = \delta^c_{a} \ c_{a} ,
\end{equation}
where $c_a$ is the original link capacity for link $a$, and $\delta^c_{a} \sim U(0.9, 1.0)$ is the random scaling factor for link $a$, assumed to remain constant over time.

\subsubsection{Experiment Settings}
The dataset is constructed under two distinct configurations for our study:
\begin{itemize}
    \item Complete OD Configuration: The OD demands are assumed to be fully known in the training and testing phases.
    \item Incomplete OD Configuration: The OD demands are assumed to be incomplete in the training process due to  missing information. The OD demand of the dataset is updated using a binary mask determined by a specified missing ratio.
\end{itemize}

For each transportation network under study, we generate a total of 1,000 samples. This sample size reflects data gathered during peak traffic periods of approximately three years, providing a robust dataset for evaluating the network's performance and resilience under varied conditions. The dataset is split into a training set, validation set, and test set randomly with a ratio of 8:1:1, respectively. 
For both configurations, all the models are optimized by the Adam optimizer \cite{loshchilov2018decoupled}  and the decay rate for momentum and squared gradient are set to 0.9, 0.999, respectively. The number of epochs and learning rate are set to 100 and 0.001, respectively. Each experimental setup is executed three times with different random seeds.

\subsubsection{Comparison Methods}
To validate the performance of HSTGSN, we have selected multiple baseline models for comparison, including predefined graph-based methods, attention-based methods, and heterogeneous graph-based methods. The descriptions of the baseline models are presented as follows:
\begin{itemize}
    \item GAT-LSTM \cite{wu2018graph}: A graph neural network that leverages attention mechanisms to focus on specific parts of a graph, enhancing node feature learning in graph-structured data.
    \item GCN-LSTM \cite{li2019hybrid}: A graph-based neural network model that utilizes convolutional techniques to aggregate information from a node's neighbors in a graph structure. 
    \item STGCN \cite{yu2017spatio}: A model that synergizes spatial graph convolutions with temporal convolutional networks to efficiently process spatio-temporal data.
    \item DCRNN \cite{li2017diffusion}: A spatio-temporal network integrating diffusion mechanism and recurrent neural network.
    \item GaAN \cite{zhang2018gaan}: A gated attention mechanism to enhance node representation learning in graph-structured data.
    \item HGT: \cite{hu2020heterogeneous}: A network employing adaptive dependent parameters to characterize the heterogeneous attention for different types of nodes and edges.
\end{itemize}

\subsubsection{Metrics}
As quantities of interest we consider two key quantities on each link: (1)   traffic flow and (2) percentage utilization. The link percentage utilization represents the ratio of link flow to link capacity, serving as a more intuitive metric to measure traffic congestion level. For comparing the predicted quantities of interest with the ground truth values, we report the mean absolute error (MAE) and the root mean square error (RMSE) as the evaluation metrics.

\subsection{Comparative Results}
We first evaluate the prediction performance of the models across all network links under complete OD configuration. The average MAE and RMSE on three networks are shown in Table \ref{tab:performance_table}. In terms of MAE, our proposed model significantly improves the dynamic user equilibrium traffic prediction accuracy on all three networks. In particular, the improvements compared to the best  baseline models, are 24.8\%, 5.4\%, and 61.4\%, respectively for the three networks. Furthermore, we performed the paired sample t-test to demonstrate the advantage of HSTGSN over other baselines from statistical perspective. As shown in Table \ref{tab:performance_table}, numbers marked with ``$\ast$" indicate that the improvement is statistically significant compared with the best baseline. In what follows, we offer  observations and interpretations based on the prediction results.
\begin{table}[htb!]
\centering
\caption{The prediction performance comparison among different models under complete OD configuration. Three networks are considered in the experiments: Sioux Falls, Anaheim, and Chicago. The symbol ``$\ast$" denotes the p-value of the t-test compared with the second-best performance is lower than 0.01.}
\label{tab:performance_table}
\begin{tabular}{cccccc}
\hline
\multirow{2}{*}{Network} & \multirow{2}{*}{Model} & \multicolumn{2}{c}{Flow} & \multicolumn{2}{c}{Link utilization} \\ \cline{3-6} 
 &  & MAE & RMSE & MAE & RMSE \\ \hline
\multirow{7}{*}{\makecell{Sioux \\Falls}} & GAT-LSTM & 78.95 & 182.04 & 3.54\% & 7.69\% \\
 & GCN-LSTM & 105.88 & 246.54 & 4.47\% & 9.25\% \\
 & STGCN & 199.70 & 392.49 & 10.44\% & 20.45\% \\
 & DCRNN & 186.36 & 354.41 & 9.90\% & 18.73\% \\
 & GaAN & 176.23 & 335.16 & 9.37\% & 17.71\% \\
 & HGT & 27.25 & 52.56 & 1.27\% & 2.41\% \\
 & Ours & $\mathbf{20.46}^{\ast}$ & $\mathbf{39.57}^{\ast}$ & $\mathbf{0.94\%}^{\ast}$ & $\mathbf{1.75\%}^{\ast}$ \\ \hline
\multirow{7}{*}{Anaheim} & GAT-LSTM & 228.36 & 984.76 & 2.84\% & 7.45\% \\
 & GCN-LSTM & 205.05 & 884.26 & 2.55\% & 6.69\% \\
 & STGCN & 173.92 & 726.33 & 2.18\% & 5.71\% \\
 & DCRNN & 171.77 & 720.56 & 2.16\% & 5.74\% \\
 & GaAN & 182.60 & 762.23 & 2.29\% & 6.00\% \\
 & HGT & 45.58 & 156.58 & 0.65\% & 2.01\% \\
 & Ours & $\mathbf{43.02}^{\ast}$ & $\mathbf{143.78}^{\ast}$ & $\mathbf{0.62\%}^{\ast}$ & $\mathbf{1.96\%}^{\ast}$ \\ \hline
\multirow{7}{*}{Chicago} & GAT-LSTM & 41.17 & 149.37 & 2.08\% & 6.29\% \\
 & GCN-LSTM & 40.95 & 148.56 & 2.07\% & 6.26\% \\
 & STGCN & 35.25 & 117.59 & 1.54\% & 4.56\% \\
 & DCRNN & 31.34 & 109.37 & 1.48\% & 4.39\% \\
 & GaAN & 31.88 & 111.71 & 1.52\% & 4.51\% \\
 & HGT & 16.60 & 65.32 & 0.86\% & 2.72\% \\
 & Ours & $\mathbf{6.40}^{\ast}$ & $\mathbf{17.35}^{\ast}$ & $\mathbf{0.18\%}^{\ast}$ & $\mathbf{0.46\%}^{\ast}$ \\ \hline
\end{tabular}
\end{table}

\begin{table*}[htb!]
\centering
\caption{The prediction performance comparison (flow,  link utilization percentage) among different models for urban transportation networks. The symbol ``$\ast$" denotes the p-value of the t-test compared with the second-best performance is lower than 0.001.}
\label{tab:performance_incomplete_table}
\begin{tabular}{cccccccccc}
\hline
\multirow{3}{*}{Network} & \multirow{3}{*}{Model} & \multicolumn{4}{c}{Missing ratio=20\%} & \multicolumn{4}{c}{Missing ratio=40\%} \\ \cline{3-10} 
 &  & \multicolumn{2}{c}{Flow} & \multicolumn{2}{c}{Link utilization} & \multicolumn{2}{c}{Flow} & \multicolumn{2}{c}{Link utilization} \\ \cline{3-10} 
 &  & MAE & RMSE & MAE & RMSE & MAE & RMSE & MAE & RMSE \\ \hline
\multirow{7}{*}{Sioux Falls} & GAT-LSTM & 168.07 & 331.93 & 4.51\% & 9.32\% & 100.59 & 291.62 & 3.90\% & 8.51\% \\
 & GCN-LSTM & 106.78 & 301.25 & 4.12\% & 8.80\% & 104.78 & 299.25 & 4.06\% & 8.77\% \\
 & STGCN & 226.17 & 424.20 & 12.20\% & 23.23\% & 231.61 & 433.67 & 12.45\% & 23.17\% \\
 & DCRNN & 241.63 & 451.77 & 12.86\% & 24.27\% & 257.35 & 481.85 & 13.83\% & 25.75\% \\
 & GaAN & 233.22 & 432.68 & 12.54\% & 23.33\% & 249.84 & 467.81 & 13.43\% & 25.00\% \\
 & HGT & 38.15 & 79.57 & 1.60\% & 2.92\% & 36.56 & 77.12 & 1.59\% & 2.98\% \\
 & Ours & $\mathbf{24.81}^{\ast}$ & $\mathbf{49.24}^{\ast}$ & $\mathbf{1.16\%}^{\ast}$ & $\mathbf{2.23\%}^{\ast}$ & $\mathbf{23.40}^{\ast}$ & $\mathbf{46.81}^{\ast}$ & $\mathbf{1.13\%}^{\ast}$ & $\mathbf{2.24\%}^{\ast}$ \\ \hline
\multirow{7}{*}{Anaheim} & GAT-LSTM & 235.05 & 970.08 & 2.88\% & 6.85\% & 247.31 & 1055.03 & 3.06\% & 7.09\% \\
 & GCN-LSTM & 219.78 & 922.07 & 2.76\% & 6.76\% & 228.76 & 975.91 & 2.83\% & 7.31\% \\
 & STGCN & 207.43 & 872.08 & 2.51\% & 6.51\% & 255.55 & 1090.20 & 3.17\% & 8.17\% \\
 & DCRNN & 221.24 & 921.08 & 2.70\% & 6.68\% & 215.16 & 917.88 & 2.67\% & 6.88\% \\
 & GaAN & 195.08 & 822.09 & 2.27\% & 6.26\% & 205.46 & 899.02 & 2.50\% & 6.62\% \\
 & HGT & 47.16 & 159.30 & 0.66\% & 2.02\% & 49.74 & 165.50 & 0.69\% & 2.04\% \\
 & Ours & $\mathbf{45.24}^{\ast}$ & $\mathbf{157.04}^{\ast}$ & $\mathbf{0.64\%}$ & $\mathbf{2.01\%}$ & $\mathbf{44.54}^{\ast}$ & $\mathbf{158.30}^{\ast}$ & $\mathbf{0.63\%}^{\ast}$ & $\mathbf{2.02\% }$\\ \hline
\multirow{7}{*}{Chicago} & GAT-LSTM & 28.05 & 100.37 & 1.44\% & 4.29\% & 28.33 & 101.38 & 1.45\% & 4.33\% \\
 & GCN-LSTM & 24.76 & 84.91 & 1.27\% & 3.93\% & 26.91 & 96.31 & 1.37\% & 4.11\% \\
 & STGCN & 24.65 & 88.20 & 1.26\% & 3.77\% & 23.79 & 85.16 & 1.21\% & 3.64\% \\
 & DCRNN & 26.40 & 92.64 & 1.35\% & 4.11\% & 23.51 & 84.15 & 1.20\% & 3.59\% \\
 & GaAN & 27.23 & 96.50 & 1.39\% & 4.20\% & 27.21 & 93.31 & 1.39\% & 4.31\% \\
 & HGT & 16.43 & 66.10 & 0.82\% & 2.57\% & 21.34 & 74.77 & 1.00\% & 3.06\% \\
 & Ours & $\mathbf{6.72}^{\ast}$ & $\mathbf{14.56}^{\ast}$ & $\mathbf{0.14\%}^{\ast}$ & $\mathbf{0.41\%}^{\ast}$ & $\mathbf{8.15}^{\ast}$ & $\mathbf{18.19}^{\ast}$ & $\mathbf{0.19\%}^{\ast}$ & $\mathbf{0.50\%}^{\ast}$ \\ \hline
\end{tabular}
\end{table*}

First, the predefined graph-based methods don't work well when applied to DUE-based traffic prediction tasks. For instance, methods like GCN-LSTM rely on predefined topologies, which may not capture dynamic spatial dependencies between source and destination nodes, thereby compromising their prediction performance. On the other hand, leveraging heterogeneous graph-based models, which employ multiple link types to facilitate feature propagation, markedly enhances the precision of DUE-based traffic flow prediction.

Second, attention-based graph methods generally perform better on larger networks than smaller networks, which may be due to the fact that they could effectively exploit valuable and latent spatial dependencies from the OD demand. However, compared with our model, the dynamic characteristics of OD demand and the spatial dependencies between source and destination nodes are neglected. The proposed HSTGSN model addresses this performance gap by introducing auxiliary OD links and adaptive edge features.

Third, the additional information from the OD links significantly improves the accuracy of traffic flow predictions. Compared with the predefined graph-based method and attention-based method, the heterogeneous graph-based approaches (HGT and our model) have smaller prediction errors across three networks. Moreover, it is noticed that the HGT underperforms our model, specifically on a larger network. To illustrate, although HGT constructed multiple link types, similar to our model, it doesn't consider the link features adaptively. This limitation may account for the degradation in performance, underscoring the importance of edge features in traffic flow prediction models. By incorporating these critical elements, the HSTGSN model consistently outperforms other spatial-temporal graph neural networks, demonstrating better performance in capturing the dynamic traffic flow.

To further assess the robustness and generalization capabilities of our model, we conducted additional experiments under an incomplete OD configuration. These experiments were conducted under two scenarios of OD demand incompleteness, with the missing ratios of 20\% and 40\%, respectively. As can be seen in the experimental results shown in Table \ref{tab:performance_incomplete_table}, our spatio-temporal encoder-decoder structure can effectively help with data imputation and OD demand reconstruction. Furthermore, the paired t-test indicates that the performance improvement of HSTGSN is statistically significant compared with the other baseline models. The results in Table \ref{tab:performance_table} and Table \ref{tab:performance_incomplete_table} indicate our model performs better than these methods consistently, underscoring the superior accuracy of our proposed model over other graph-based methods.

\subsection{Ablation Study}
To evaluate the effectiveness of each component within the HSTGSN framework, we conducted an ablation study by  excluding various components of the model  one at a time. In particular, we considered three cases: (1) ``w/o link feat", which is HSTGSN without road link features, thereby omitting the consideration of link-specific variations; (2) ``w/o OD link", which is HSTGSN without any OD link, forming a  homogeneous graph with only physical links; and (3) ``w/o Adaptive", which is HSTGSN after removing the learnable parameters in the attention calculation for the spatial encoder-decoder.

\begin{figure}[hbt!]
  \includegraphics[width=\linewidth]{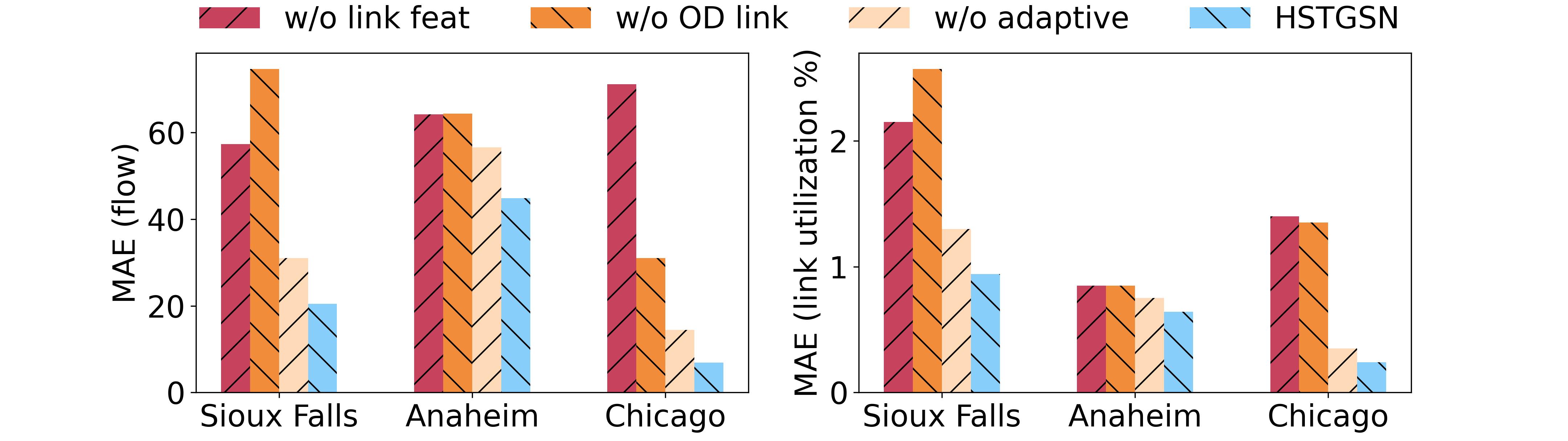}
  \caption{Ablation study of different variants of HSTGSN. Three variations of HSTGSN are included: ``w/o link feat": model maintaining OD link but removing road link features; ``w/o OD link": model removing OD link; ``w/o Adaptive": model removing adaptive attention mechanism. }
  \label{fig:ablation}
\end{figure}

The comparative analysis, illustrated in Figure \ref{fig:ablation}, shows a degradation in performance across all modified cases compared to the complete model,  underscoring the important contribution of each of the those components. Notably, it can be seen that the addition of OD links plays a very significant role, as their removal result in large error increases in flow and link utilization predictions. This is because the integration of OD links in HSTGSN alleviate the need for numerous embedding propagation steps through road links, leading to improved model efficiency. Moreover, removing link features from the OD link compromises model performance, indicating the significance of the link features in traffic flow prediction.  Furthermore, removing adaptive parameters from the model slightly increases prediction errors, indicating that the adaptive representation of link features can further enhance  the learning of dynamic traffic flow distribution.

\section{Conclusion}
\label{sec:conslusion}
Traffic flow prediction provides critical insights for urban planning, traffic management, and the advancement of intelligent transportation systems. In this study, we developed a novel heterogeneous spatio-temporal graph sequence network (HSTGSN) for DUE-based traffic assignment. Compared with existing studies, which are primarily node-based traffic flow prediction relying on sensor networks, our approach extends the applicability of traffic flow prediction to every link in the region. This integration facilitates link-based traffic flow forecasting across the entire network, including areas beyond sensor locations. HSTGSN consists of multiple link types, namely roadway links and OD links, forming a heterogeneous graph, and uses an adaptive attention mechanism to construct a sequence of spatial-temporal encoders and decoders. When tested extensively on urban transportation networks, our model not only accurately captures the spatial-temporal relationships between traffic flows and origin-destination demands but also outperforms existing models in predictive accuracy and generalization capabilities across various configurations. Our proposed model can significantly enhances traffic flow analysis, enabling an urban transportation system management that is  accurate and adaptable, and yet computationally affordable for use by transportation engineers and urban planners. 

\section*{Acknowledgments}
This material is based in part upon work partially supported by the National Science Foundation under Grant No. CMMI-1752302

\bibliographystyle{IEEEtran}
\bibliography{reference}

\end{document}